\documentclass[runningheads,a4paper]{llncs}

\usepackage{amssymb}
\usepackage{graphicx}
\usepackage{times}
\usepackage{latexsym}
\usepackage{float}
\usepackage[normalem]{ulem}
\usepackage[T1]{fontenc}
\usepackage{graphicx}
\usepackage{url}
\usepackage{subcaption}
\urldef{\mailsa}\path|{souvick.ghosh}@rutgers.edu|
\urldef{\mailsb}\path|{satanu.ghosh.94}@gmail.com|
\urldef{\mailsc}\path|dipankar.dipnil2005}@gmail.com|

\begin{document}

\mainmatter  

\title{Sentiment Identification in Code-Mixed Social Media Text}


%
%

\author{Souvick Ghosh\textsuperscript{1} \and Satanu Ghosh\textsuperscript{2}\and Dipankar Das\textsuperscript{3} \\}



\institute{\textsuperscript{1}Rutgers University, New Brunswick, NJ 08901, US\\
\mailsa\\
\textsuperscript{2}MAKAUT, Kolkata 700064, WB, India\\
\mailsb\\
\textsuperscript{3}Jadavpur University, Kolkata 700032, WB, India\\
\mailsc\\}

\maketitle

\begin{abstract}
Sentiment analysis is the Natural Language Processing (NLP) task dealing with the detection and classification of sentiments in texts\nocite{Balahur:13}. While some tasks deal with identifying presence of sentiment in text (Subjectivity analysis), other tasks aim at determining the polarity of the text categorizing them as \textit{positive}, \textit{negative} and \textit{neutral}. Whenever there is presence of sentiment in text, it has a source (people, group of people or any entity) and the sentiment is directed towards some entity, object, event or person. Sentiment analysis tasks aim to determine the subject, the target and the polarity or valence of the sentiment.
In our work, we try to automatically extract sentiment (positive or negative) from Facebook posts using a machine learning approach. While some works have been done in code-mixed social media data and in sentiment analysis separately, our work is the first attempt (as of now) which aims at performing sentiment analysis of code-mixed social media text. We have used extensive pre-processing to remove noise from raw text. Multilayer Perceptron model has been used to determine the polarity of the sentiment. We have also developed the corpus for this task by manually labelling Facebook posts with their associated sentiments. 
\end{abstract}

\section{Introduction}

Sentiment analysis  - of social media in particular - has become a popular area of research in present times. The massive proliferation of social media has been a catalyst in this regard. A culture shift can be noticed where the users comfortably and candidly express their emotions, opinions or sentiments online. This has encouraged the researchers to analyze and study the presence of sentiments from social media. 

Extraction of sentiment from social media – like Facebook  or microposts like Twitter  – can serve a myriad of purposes. These texts often express opinion about a variety of topics. It can be the appraisal of the user about certain products or incidents, the state of mind of the speaker or any intended emotional communication that he may want to have with potential readers. User reviews on e-commerce sites, opinions on web blogs, tweets\footnote{twitter.com} and Facebook\footnote{www.facebook.com} posts, can be mined for assessing polarity of opinion. Businesses use the power of text analytics behind their data mining technology. Sentiment analysis helps businesses in advertising, marketing and making business decisions for better customer satisfaction. Organizations can determine public opinion about their products and services. Similarly, consumers can use sentiment analysis while researching products prior to purchase. It can also be used to investigate the web for forecasting electoral results (by evaluating voter sentiment) and track political preferences. Recently, social media analysis has been used extensively to identify cyber-bullying prevalent in the web space \cite{nahar2012sentiment}.

Although we have come across various tasks conducted on multilingual texts, the task of sentiment analysis, in particular, has not been explored for multilingual code-mixed texts. This type of text differs significantly from traditional English texts and needs to be processed differently. However, different forms of texts require different methods for sentiment analysis. For example, if we look at sentiments in scientific papers, it is hedged and indirect while the sentiments are more direct in movie or product reviews. Traditional texts like reviews and newspaper are structured and follow a definite pattern. Also, the writing is more formal and composed. Social media texts on the other hand are largely informal. They are concise and informal with several linguistic differences. 

In our work, we have used code-mixed social media data which have been collected from Facebook post. The text is informal and conversational in accordance with social media characteristics. It is mostly bilingual though the presence of three languages in a single post is not entirely uncommon in our data. Initially, we pre-process the text to normalize the irregular words. We also remove noise from the text prior to processing it and translate the abbreviations to regular words wherever applicable. We label the posts with their respective part-of-speech tags. Traditionally, sentiment classifiers show improvements by using part-of-speech features. We make use of various word-level, dictionary-based and stylistics features relevant to social media text to classify the sentiment as subjective or objective. Subjective posts are further categorized as positive or negative in polarity. We use various machine learning algorithms for our final classification. Artificial neural network model performs best in our experiments.

The remainder of this paper is structured as follows: Section 2 gives an overview of the background and related work. In Section 3, we present the dataset. The working model for our system is described in Section 4. We describe in detail the pre-processing and feature selection used to build the classification models. In Section 5, we present the results obtained using different combinations of features. We evaluate the performance of various machine learning models that we used in our experimentation. Section 6 summarizes the main findings of this work and sketches the lines for future work.

\section{Related Work}

Research regarding emotion and mood analysis in text – is becoming more common recently, in part due to the availability of new sources of subjective information on the web. The work of ~\cite{Ortony:87} was one of the very first in the area of sentiment classification. They focused on the actual taxonomy and isolation of terms with an emotional connotation.

Identifying the semantic polarity (positive vs. negative connotation) of words has been done using different approaches. Some of the works (knowledge-based) explicitly attempted to find features indicating that subjective language is being used. ~\cite{Hatzivassiloglou:97} made use of corpus statistics, ~\cite{Wiebe:00} used linguistic tools such as WordNet ~\cite{Kamps:04}, and ~\cite{Liu:03} used lexicon-based classifier. ~\cite{Turney:02} work on classification of reviews was based on using an unsupervised learning technique. They found the mutual information between document phrases and the words like “excellent” and “poor”. The mutual information was computed using statistics gathered by a search engine. In their work on automatic classification of sentiment in online domains, ~\cite{Pang:02} evaluated the performance of different classifiers on movie reviews. They demonstrated that that standard machine learning techniques outperform human-produced baselines.

Typically, methods for sentiment analysis produce lists of words with polarity values assigned to each of them. This method has been successfully employed for applications such as product review analysis and opinion mining ~\cite{Das:01,Dave:03,Grefenstette:04,Pang:02,Nasukawa:03,Turney:03,Esuli:06}. ~\cite{Holzman:03} reported high accuracy in classifying emotions in online chat conversations by using the phonemes extracted from a voice-reconstruction of the conversations. ~\cite{Rubin:04} investigated discriminating terms for emotion detection in short text while ~\cite{Read:04} described a system for identifying affect in short fiction stories, using the statistical association level between words in the text and a set of keywords. In another work, ~\cite{Read:05} used distant supervision to build the corpus.

There has been some work by researchers in the area of phrase level and sentence level sentiment classification ~\cite{Wilson:05} and on analyzing blog posts ~\cite{Mishne:05}. ~\cite{Wilson:05} determined whether an expression is neutral or polar and then disambiguated the polarity of the polar expressions. With this approach, their system was able to automatically identify the contextual polarity for a large subset of sentiment expressions.

Sentiment analysis of social media text has received a lot of interest from the research community in the recent years with the rise to prominence of Facebook and Twitter. ~\cite{Ding:08} used context-dependent sentiment words in their work and ~\cite{Tan:08} suggested combining learning-based and lexicon-based techniques using a centroid classifier. ~\cite{Go:09} used positive and negative emoticons to classify tweet polarity. They showed that machine learning algorithms (Naive Bayes, Maximum Entropy, and SVM) have accuracy above 80\% when trained with emoticon data. ~\cite{Pak:10} showed how to automatically collect a corpus for sentiment analysis and opinion mining purposes. They concluded that authors use syntactic structures to describe emotions or state facts and some POS-tags may be strong indicators of emotional text. They obtained best results using Naive Bayes classifier that uses N-gram and POS-tags as features. ~\cite{Diakopoulos:10} used crowdsourcing techniques to manually rate polarity in Twitter posts. In their work, ~\cite{De:12} classified human affective states from posts shared on Twitter. ~\cite{Wang:12} highlighted the suitability of Support Vector Machine or Naive Bayes for different domains. Our approach is similar to that of ~\cite{Zhang:11} who presented the idea of ternary classification system (positive, negative and neutral). They used target words bearing sentiment and supervised learning for classification. We also use some techniques for noise reduction which was inspired by ~\cite{Hu:13}. They proposed building a sophisticated feature space to handle noisy and short messages in their work on Twitter sentiment analysis.

\section{Dataset}

A recent shared task was conducted by Twelfth International Conference on Natural Language Processing (ICON-2015)\footnote{http://ltrc.iiit.ac.in/icon2015/contests.php} , for part-of-speech tagging of transliterated social media text. For the shared task in that corpus, data was collected from Bengali-English Facebook chat groups. The Facebook posts are in mixed English-Bengali and English-Hindi – and have been obtained from the “JU Confession” Facebook group, which contains posts in English-Bengali with few Hindi words in some cases.
We have modified the ICON Shared Task Corpora for our work on sentiment analysis. The dataset contains three languages – Bengali, Hindi and English. The data set contains 882 posts in total. The statistics for the dataset have been presented in Table \ref{table:1}.

\begin{table}[!htbp]
\small
\centering
\begin{tabular}{|p{3.8cm}|p{3.8cm}|p{3.8cm}|}
\hline \bf Language Tags & \bf Number Of Words Present & \bf Percentage Of Corpus \\ \hline
English (En) & 9988 & 52.72 \\
Bengali (Bn) & 8330 & 43.97 \\
Hindi (Hi) & 626 & 3.3 \\
\hline
\end{tabular}
\caption{\label{font-table} Statistics of the Corpus.}
\label{table:1}
\end{table}

The purpose of the implementation is to be able to automatically classify a post as a positive or negative tweet sentiment wise. The classifier needs to be trained and to do that we needed a list of manually classified posts. We used 2 annotators to classify the posts into three categories – positive, negative or neutral.

We have calculated Kappa co-efficient to measure the inter-annotator agreement. Kappa co-efficient is a reliable and robust measure to measure the agreement between two users. It takes into account the agreement occurring by chance and hence, is more useful than percent agreement calculation.

\begin{table}
\small
\centering
\begin{tabular}{|p{2.25cm}|p{2cm}|p{2cm}|p{2cm}|p{2cm}|}
\hline 
\bf Annotator 1 & \multicolumn{3}{c|}{\bf Annotator 2} & \\ 
\hline
& Positive & Neutral & Negative & Total \\
\hline
Positive & 200 & 146 & 13 & 359 \\
Neutral & 46 & 268 & 26 & 340 \\
Negative & 6 & 80 & 97 & 183 \\
\hline
Total & 252 & 494 & 136 & \\
\hline
\end{tabular}
\caption{\label{font-table} Inter-Annotator Agreement.}
\label{table:2}
\end{table}

For the above data, po is 0.641 and pe is 0.3642, therefore giving a Kappa co-efficient of 0.4354. Because the Kappa measure is low, so we have obtained the instances where the annotators are unanimous about the sentiment polarity. There are a total of 565 such instances. We have used these posts for our sentiment polarity classification.

\section{System Description}

\begin{figure*}[!htpb]
  \includegraphics[width=\linewidth]{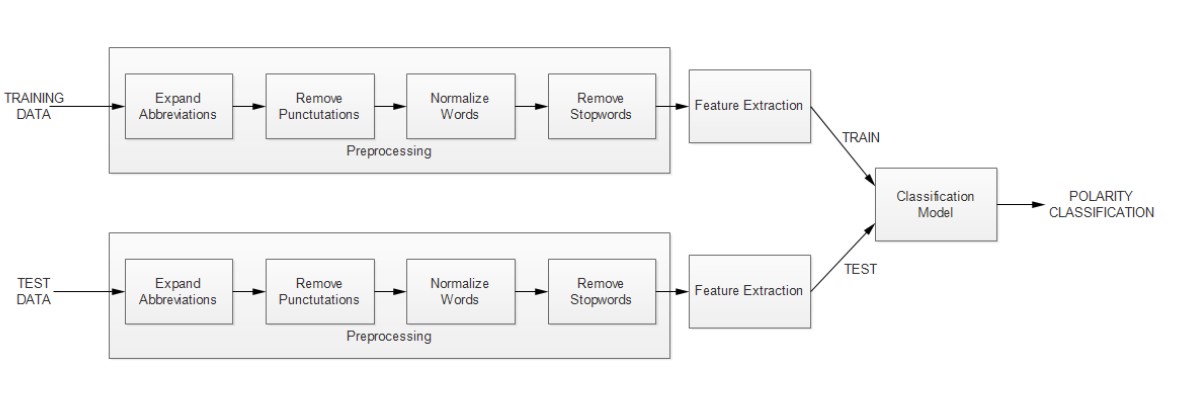}
  \caption{\label{font-figure} Overview of the System Architecture.}
  \label{fig:1}
\end{figure*}

The process of sentiment analysis can be divided into three major parts : pre-processing of raw posts, feature identification and extraction and finally, the classification of sentiment as positive, neutral or negative. The steps have been discussed in sequential order.

\subsection{Pre-processing of the Facebook posts}
The following steps were performed to pre-process the raw posts prior to feature extraction.

\begin{enumerate}
\item \textbf{Expansion of Abbreviations}

As social media text is often non-traditional and informal in nature, the posts had to be pre-processed initially to remove noise. We have used an abbreviation list to normalize all the words that were abbreviated. For example, \textit{btw} was replaced by \textit{“by the way”}, \textit{clg} by \textit{“college”}, \textit{hw} by \textit{“how”} and so on.

\item \textbf{Removal of Punctuations}

Before processing the post any further, we remove all punctuations from the text. Mostly social media texts contains a lot of punctuations and their usage is often arbitrary in nature, not adhering to grammatical norms. To compound the problem further, punctuations like stop, question mark and exclamation marks are often used multiple times in succession. By removing all the punctuations, we try to make our text as noiseless as possible. We keep a record of the number of different punctuations in the text which has been used as a feature for classification.

\item \textbf{Removal of Multiple Character Repetitions}

It is often found in social media text that certain characters are repeated more than once. These non-conformational spellings are very hard to deal with as they cannot be successfully matched to any dictionary. For example, \textit{lol} (abbreviated form of \textit{laughing out loud})can be written as \textit{loool}, \textit{looool} or \textit{loooooool}. We use pre-processing in order to reduce all these occurrences to lool. Any character which occurs more than two times in a row is replaced by two occurrences of the same character. Some other examples are \textit{ahhhh} (reduced to \textit{ahh}) and \textit{uhhhh} (reduced to \textit{uhh}). However, we maintain a record of the number of repetitions as this could be used by the author in specific situations to reflect sentiment.

\end{enumerate}
\subsection{Feature Extraction}

In our work, we used the following features to train our machine learning model.

\begin{enumerate}

\item \textbf{Number Of Word Matches With Sentiwordnet (SWN):}
We have used SentiWordNet\footnote{http://sentiwordnet.isti.cnr.it/} as one of the sentiment resources. SWN is a lexical resource for sentiment analysis. It assigns three sentiment scores – positivity, negativity and objectivity to each synset of WordNet. So, a given word can have a positive or negative score or both. We have extracted all the positive and negative words from SWN. The final list contains 17027 positive words and 17992 negative words. For a given data instance or sentence, we find if the normalized words are a match with any words in these two lists. In a sentence we count the number of words which matches with the positive word list and the number of words which matches with the negative word list and the assign the difference between the positive and negative word count as a feature.

\item \textbf{Number Of Word Matches With Opinion Lexicon (OL):}
Similar to SentiWordNet, Opinion Lexicon\footnote{https://www.cs.uic.edu/~liub/FBS/sentiment-analysis.html}  is another lexical resource for sentiment analysis. It contains a list of positive and negative opinion words or sentiment words for English. There is a total of 2006 positive words and 4783 negative words. We find the number of matches to both the lists and the difference is taken as our second feature.

\item \textbf{Number Of Word Matches With English Sentiment Words (ESW):}
We have collected a list of positive and negative words from the internet for sentiment classification. We hand-labeled a few words in the training data as root words which depicted emotion. Using bootstrapping, we expanded this list of words. It contains 3075 positive words and 4003 negative words. This list concentrates more on the words which appear in social media context. Similar to the previous two features, we find the number of matches to the positive and negative lists and the difference between the two is considered as our third feature. 

\item \textbf{Number Of Word Matches With Bengali Sentiment Words (BSW):}
This list was developed to tackle the presence of sentiment in Bengali words. As we are dealing with multilingual text, it was essential to develop this list for Bengali. Das and colleagues~\cite{Das:10,Das:11,Das:12} developed SentiWordNet for Indian Languages. However, this list contained words in Bengali (or Brahmic) scripts. As we are dealing with transliterated text, this wordlist required transliteration to English. Finally, we developed a positive and a negative wordlist for transliterated Bengali words. The number of words in the positive wordlist is 1778 while the negative wordlist contains 3713 words. The difference in number of matches to both the lists is considered as our next feature.

\item \textbf{Number Of Colloquial Bengali Sentiment Words (CBW):}
We have created this list for Bengali words which often appear in social media text. It must be noted that Bengali Sentiment Words developed previously is more formal in nature and therefore, not sufficient for identifying colloquial words which appear in Facebook posts or Twitter texts. For example, words like \textit{jata} (hopeless), \textit{hebby} (excellent), \textit{phot} (get lost) are not captured by Bengali Sentiment Words. We create two lists – positive and negative wordlists - tries to incorporate all such words which may indicate the presence of sentiment in the text. The number of matches to both the lists is determined and the difference is assigned as feature.

\item \textbf{Density Of Curse Or Bad Words (CW):}
We have used a list of curse words (words which are used as bad words in majority instances) developed by ~\cite{Huang:14} in their work on cyberbullying. In their work, the authors collected 713 curse words (e.g. `asshole', `bitch' etc.) and hieroglyphs (such as `5hit', `@ss' etc.) based on online resources. We have used this list to find out all the words which have been used with a negative sentiment.

\item \textbf{Part-Of-Speech Tags (POS):}
All the posts were tagged manually for parts-of-speech information. It has been noted that words belonging to certain part-of-speech tags (like JJ, RB and JJ-RB) are usually used to express sentiment. These part-of-speech tags can be considered as features to detect presence of sentiment in commonly occurring unigram and bigrams in the training data. 

\item \textbf{Number Of All Uppercase Words (UW):}
Based on the findings of ~\cite{Dadvar:13}, capital letters can represent shouting or strong opinion in online chats and posts. We have identified the number of words in a post which are written in all capital letters. This is used as a feature to detect the presence of emotion or sentiment in online settings.

\item \textbf{Density Of Exclamation Points (E):}
Just like the uppercase letters, exclamation points also stand as emotional comments. To identify strong emotions in social media context, we chose the number of exclamation points as a feature for our model. The number of exclamation points is normalized by the number of words present in the text.

\item \textbf{Density Of Question Marks (Q):}
Similar to the last feature, multiple question marks in the text can denote surprise, excitement or agitation of the user. We chose the number of question marks as our next feature. The number of question marks is normalized by the number of words present in the text.

\item \textbf{Number Of Character Repetitions In A Word (R):}
It is often observed that users tend to repeat a number of characters – vowels or consonants – to stress their opinion in social media conversations. Words like \textit{loool}, \textit{lolzzzz}, \textit{ufffff}, \textit{ahaaa}, \textit{greaaat} are quite common in social media texts. While we reduce all such words during our pre-processing step, we have also maintained a record of all such occurrences. These repetitions are often indicative of sentiment and we use it as one of our feature.

\item \textbf{Frequency Of Code Switches (CS):}
As we are dealing with multilingual texts, we have considered the frequency of code switching as one of our features. It is often observed that the writer shifts language to clarify his opinion. We have tried to exploit this social and communication needs for this language shifting to determine the presence of sentiment. This frequency (number of language switching points) is normalized by the number of words in a particular post.

\item \textbf{Number Of Smiley Matches (S1 And S2):}
Smileys are quite prevalent in social media text and often form a primary way of expressing emotion. We have created two resources for identifying smiley in text. The first one contains 269 positive smileys and 170 negative smileys. The second list contains 243 smileys. We found the number of matches to both the lists and used it as a feature.

\end{enumerate}

\subsection{Classification of Sentiment Polarity}
We obtain results for the 565 posts for which both the annotators agreed on the polarity. We use 70\% of the dataset for training and 30\% for testing purposes.We split the dataset using 400 posts for training and 165 posts for testing.

We use the machine learning software WEKA\footnote{http://www.cs.waikato.ac.nz/ml/weka/downloading.html} ~\cite{Hall:09}. We combine the above features to form a feature set and employ a number of machine learning algorithms for classification. The best results were produced by Multilayer Perceptron model. This classifier uses back propagation to classify instances into three categories – \textit{positive}, \textit{negative} and \textit{neutral}. The nodes in this network are all sigmoid. The learning rate and momentum rate for the back propagation algorithm was kept at 0.3 and 0.2 respectively. The number of epochs was set to 500 and the random number generator was seeded using value 0.

Individually, none of the features was able to detect positive or negative instances in citation. This is due to the biasness of the system. We perform feature analysis by removing one feature at a time to determine if any feature is more important than the other. We also check by adding one feature group at a time. The classification confidence score from WEKA and the number of matches to our citation specific lexicon is used to develop a post-processing algorithm.

\section{Results and Observations}
For feature analysis, we have grouped the different kind of features and obtained the impact of each group in classification. We have grouped the word (or dictionary) based features into Group 1 (G1), syntactic features into Group 2 (G2) and the style based features into Group 3 (G3).
\newline \textit{G1: SWN + OL + ESW + BSW + CBW + CW + S}
\newline \textit{G2: POS}
\newline \textit{G3: UW + E + Q + R + CS}

\begin{table}[!htpb]
\small
\centering
\begin{tabular}{|p{2.4cm}|p{2.4cm}|p{2.4cm}|p{2.4cm}|}
\hline \bf Feature added & \bf Correct classifications & \bf Incorrect classifications & \bf Accuracy \\ \hline
G1 & 110 & 55 & 0.667 \\
\bf G1 + G2 & \bf 113 & \bf 52 & \bf 0.685 \\
G1 + G2 + G3 & 101 & 64 & 0.612 \\
\hline
\end{tabular}
\caption{\label{font-table} Impact of Adding Each Feature Iteratively To the Last.}
\label{table:3}
\end{table}

From Table \ref{table:3} it is evident that word based features (Group 1) and syntactic features (Group 2) produce the best results collectively. The accuracy decreases when we include the style based features for classification.
\newline

Table \ref{table:4} serves to highlight the impact of individual features in classification. At each turn, we eliminate one of the features while keeping all the other features. The accuracy suffers the maximum on elimination of POS (JJ, RB and RB\_JJ) features and the polar smiley list. Elimination of all the style based features (UW, E, Q, R and CS) shows improvement in accuracy. This is in accordance to our findings in Table \ref{table:3}. Elimination of SWN also improves accuracy. Removing BSW – which comprises of conformational (or traditional) Bengali words – do not affect accuracy proving the fact that social media text requires tailor-made resources.

\begin{table}[!htpb]
\small
\centering
\begin{tabular}{|p{2.4cm}|p{2.4cm}|p{2.4cm}|p{2.4cm}|}
\hline 
\bf Feature Eliminated & \bf Correct classifications & \bf Incorrect classifications & \bf Accuracy \\
\hline
None & 104 & 61 & 0.630 \\
SWN & 109 & 56 & 0.661 \\
OL & 103 & 62 & 0.624 \\
ESW & 102 & 63 & 0.618 \\
BSW & 104 & 61 & 0.630 \\
CBW & 101 & 64 & 0.612 \\
S & 105 & 60 & 0.636 \\
\bf POS & \bf 100 & \bf 65 & \bf 0.606 \\
UW & 110 & 55 & 0.667 \\
E & 107 & 58 & 0.649 \\
Q & 105 & 60 & 0.636 \\
R & 107 & 58 & 0.649 \\
CS & 106 & 59 & 0.642 \\
\bf S1 & \bf 100 & \bf 65 & \bf 0.606 \\
S2 & 106 & 59 & 0.642 \\
\hline
\end{tabular}
\caption{\label{font-table} Impact Of Each Feature Calculated By Eliminating One at A Time.}
\label{table:4}
\end{table}

Table \ref{table:5} shows the confusion matrix for the polarity classification (using word based and semantic features). The precision, recall and f-measure of the supervised and baseline systems are compared in Table \ref{table:6}.

\begin{table}[!htpb]
\small
\centering
\begin{tabular}{|p{2.4cm}|p{2.4cm}|p{2.4cm}|p{2.4cm}|}
\hline 
 & \bf Positive & \bf Neutral & \bf Negative \\ 
\hline
\bf Positive & 25 & 23 & 4 \\
\bf Neutral & 10 & 78 & 3 \\
\bf Negative & 4 & 8 & 10 \\
\hline
\end{tabular}
\caption{\label{font-table} Confusion Matrix for Classification.}
\label{table:5}
\end{table}

\begin{table}[!htpb]
\small
\centering
\begin{tabular}{|p{2.4cm}|p{2.4cm}|p{2.4cm}|p{2.4cm}|}
\hline 
 & \bf Precision & \bf Recall & \bf F-measure \\
\hline
\bf Class Positive & 0.641 & 0.481 & 0.55 \\
\bf Class Neutral & 0.716 & 0.857 & 0.78 \\
\bf Class Negative & 0.588 & 0.455 & 0.513 \\
\hline
\end{tabular}
\caption{\label{font-table} Precision, Recall and F-measure.}
\label{table:6}
\end{table}

If we consider the baseline model to contain all the instances of neutral polarity, then we can achieve an accuracy of 55.2\%. Our best performing system shows an accuracy of 68.5\%. So we can see that our supervised system shows improvement over the baseline model. However, the learning algorithm was slightly biased towards neutral classification which is evident from the confusion matrix. Most of the errors are due to positive and negative citations being identified as neutral.

In future works, we will need to fine tune our classification features so that the system can identify positive and negative citations more efficiently. Also using a larger dataset to train the system would eliminate the bias towards neutral classification of polarity.

\section{Conclusion and Future Work}

As per our knowledge, there exists no sentiment classifier for code-mixed social media text. We have performed a machine learning based sentiment classification of Facebook posts. The polarity of each post has been classified as \textit{positive}, \textit{negative} and \textit{neutral}. As there has not been any similar work before, we had to create a dataset of our own. Two human annotators classified the polarity of each post. Due to the inherent complexity of social media text, use of arbitrary emoticons and presence of sarcasm, the agreement between the human annotators was quite low with a Kappa co-efficient of 0.4354. Although the entire dataset consists of 882 posts, we have used only 565 posts where the annotators were unanimous about the polarity of underlying sentiment. We used word-based, semantic and style-based features for classification. The best result was obtained using a combination of word-based and semantic features with an accuracy of 68.5\%.

As our dataset is relatively small, we would like to create a larger dataset in future. Sentiment annotation can also be done using distant supervision based on the presence of emoticons. However, such an approach can lead to noisy dataset. Creating a gold standard for all future tasks is a priority for us. In this work, we have not focused on detection of sarcasm in text. Also, we have not handled negation in data. We would like to concentrate on dealing with these issues in our next work. Apart from that, sentiment classification can be further improved by better handling comparisons and by detecting sentiment targeted towards an entity in particular. Handling of context switches is also important. Developing a real time accurate sentiment classifier model is the ultimate goal which we strive to achieve in future.
\bibliography{cicling}

\begin{thebibliography}{10}
\providecommand{\url}[1]{\texttt{#1}}
\providecommand{\urlprefix}{URL }

\bibitem{Balahur:13}
Balahur, A.: Sentiment analysis in social media texts. In: 4th workshop on
  Computational Approaches to Subjectivity, Sentiment and Social Media
  Analysis, pp. 120--128 (2013)

\bibitem{Dadvar:13}
Dadvar, M., Trieschnigg, D., Ordelman, R., de~Jong, F.: Improving cyberbullying
  detection with user context. In: Advances in Information Retrieval, pp.
  693--696. Springer (2013)

\bibitem{Das:10}
Das, A., Bandyopadhyay, S.: Sentiwordnet for indian languages. Asian Federation
  for Natural Language Processing, China pp. 56--63 (2010)

\bibitem{Das:11}
Das, A., Bandyopadhyay, S.: Dr sentiment knows everything! In: Proceedings of
  the 49th annual meeting of the association for computational linguistics:
  human language technologies: systems demonstrations. pp. 50--55. Association
  for Computational Linguistics (2011)

\bibitem{Das:12}
Das, A., Gamb{\"a}ck, B.: Sentimantics: conceptual spaces for lexical sentiment
  polarity representation with contextuality. In: Proceedings of the 3rd
  Workshop in Computational Approaches to Subjectivity and Sentiment Analysis.
  pp. 38--46. Association for Computational Linguistics (2012)

\bibitem{Das:01}
Das, S.R., Chen, M.Y.: Yahoo! for amazon: Sentiment parsing from small talk on
  the web. For Amazon: Sentiment Parsing from Small Talk on the Web (August 5,
  2001). EFA  (2001)

\bibitem{Dave:03}
Dave, K., Lawrence, S., Pennock, D.M.: Mining the peanut gallery: Opinion
  extraction and semantic classification of product reviews. In: Proceedings of
  the 12th international conference on World Wide Web. pp. 519--528. ACM (2003)

\bibitem{De:12}
De~Choudhury, M., Gamon, M., Counts, S.: Happy, nervous or surprised?
  classification of human affective states in social media. In: ICWSM (2012)

\bibitem{Diakopoulos:10}
Diakopoulos, N.A., Shamma, D.A.: Characterizing debate performance via
  aggregated twitter sentiment. In: Proceedings of the SIGCHI Conference on
  Human Factors in Computing Systems. pp. 1195--1198. ACM (2010)

\bibitem{Ding:08}
Ding, X., Liu, B., Yu, P.S.: A holistic lexicon-based approach to opinion
  mining. In: Proceedings of the 2008 International Conference on Web Search
  and Data Mining. pp. 231--240. ACM (2008)

\bibitem{Esuli:06}
Esuli, A., Sebastiani, F.: Sentiwordnet: A publicly available lexical resource
  for opinion mining. In: Proceedings of LREC. vol.~6, pp. 417--422. Citeseer
  (2006)

\bibitem{Go:09}
Go, A., Bhayani, R., Huang, L.: Twitter sentiment classification using distant
  supervision. CS224N Project Report, Stanford  1, ~12 (2009)

\bibitem{Grefenstette:04}
Grefenstette, G., Qu, Y., Shanahan, J.G., Evans, D.A.: Coupling niche browsers
  and affect analysis for an opinion mining application. In: Coupling
  approaches, coupling media and coupling languages for information retrieval.
  pp. 186--194. LE CENTRE DE HAUTES ETUDES INTERNATIONALES D'INFORMATIQUE
  DOCUMENTAIRE (2004)

\bibitem{Hall:09}
Hall, M., Frank, E., Holmes, G., Pfahringer, B., Reutemann, P., Witten, I.H.:
  The weka data mining software: an update. ACM SIGKDD explorations newsletter
  11(1),  10--18 (2009)

\bibitem{Hatzivassiloglou:97}
Hatzivassiloglou, V., McKeown, K.R.: Predicting the semantic orientation of
  adjectives. In: Proceedings of the 35th annual meeting of the association for
  computational linguistics and eighth conference of the european chapter of
  the association for computational linguistics. pp. 174--181. Association for
  Computational Linguistics (1997)

\bibitem{Holzman:03}
Holzman, L.E., Pottenger, W.M.: Classification of emotions in internet chat: An
  application of machine learning using speech phonemes. Retrieved November
  27(2011), ~50 (2003)

\bibitem{Hu:13}
Hu, X., Tang, L., Tang, J., Liu, H.: Exploiting social relations for sentiment
  analysis in microblogging. In: Proceedings of the sixth ACM international
  conference on Web search and data mining. pp. 537--546. ACM (2013)

\bibitem{Huang:14}
Huang, Q., Singh, V.K., Atrey, P.K.: Cyber bullying detection using social and
  textual analysis. In: Proceedings of the 3rd International Workshop on
  Socially-Aware Multimedia. pp. 3--6. ACM (2014)

\bibitem{Kamps:04}
Kamps, J., Marx, M., Mokken, R.J., Rijke, M.d., et~al.: Using wordnet to
  measure semantic orientations of adjectives  (2004)

\bibitem{Liu:03}
Liu, H., Lieberman, H., Selker, T.: A model of textual affect sensing using
  real-world knowledge. In: Proceedings of the 8th international conference on
  Intelligent user interfaces. pp. 125--132. ACM (2003)

\bibitem{Mishne:05}
Mishne, G., et~al.: Experiments with mood classification in blog posts. In:
  Proceedings of ACM SIGIR 2005 workshop on stylistic analysis of text for
  information access. vol.~19, pp. 321--327. Citeseer (2005)

\bibitem{nahar2012sentiment}
Nahar, V., Unankard, S., Li, X., Pang, C.: Sentiment analysis for effective
  detection of cyber bullying. In: Asia-Pacific Web Conference. pp. 767--774.
  Springer (2012)

\bibitem{Nasukawa:03}
Nasukawa, T., Yi, J.: Sentiment analysis: Capturing favorability using natural
  language processing. In: Proceedings of the 2nd international conference on
  Knowledge capture. pp. 70--77. ACM (2003)

\bibitem{Ortony:87}
Ortony, A., Clore, G.L., Foss, M.A.: The referential structure of the affective
  lexicon. Cognitive science  11(3),  341--364 (1987)

\bibitem{Pak:10}
Pak, A., Paroubek, P.: Twitter as a corpus for sentiment analysis and opinion
  mining. In: LREc. vol.~10, pp. 1320--1326 (2010)

\bibitem{Pang:02}
Pang, B., Lee, L., Vaithyanathan, S.: Thumbs up?: sentiment classification
  using machine learning techniques. In: Proceedings of the ACL-02 conference
  on Empirical methods in natural language processing-Volume 10. pp. 79--86.
  Association for Computational Linguistics (2002)

\bibitem{Read:04}
Read, J.: Recognising affect in text using pointwise-mutual information.
  Unpublished M. Sc. Dissertation, University of Sussex, UK  (2004)

\bibitem{Read:05}
Read, J.: Using emoticons to reduce dependency in machine learning techniques
  for sentiment classification. In: Proceedings of the ACL student research
  workshop. pp. 43--48. Association for Computational Linguistics (2005)

\bibitem{Rubin:04}
Rubin, V.L., Stanton, J.M., Liddy, E.D.: Discerning emotions in texts. In: The
  AAAI Symposium on Exploring Attitude and Affect in Text (AAAI-EAAT) (2004)

\bibitem{Tan:08}
Tan, S., Wang, Y., Cheng, X.: Combining learn-based and lexicon-based
  techniques for sentiment detection without using labeled examples. In:
  Proceedings of the 31st annual international ACM SIGIR conference on Research
  and development in information retrieval. pp. 743--744. ACM (2008)

\bibitem{Turney:02}
Turney, P., Littman, M.L.: Unsupervised learning of semantic orientation from a
  hundred-billion-word corpus  (2002)

\bibitem{Turney:03}
Turney, P.D., Littman, M.L.: Measuring praise and criticism: Inference of
  semantic orientation from association. ACM Transactions on Information
  Systems (TOIS)  21(4),  315--346 (2003)

\bibitem{Wang:12}
Wang, S., Manning, C.D.: Baselines and bigrams: Simple, good sentiment and
  topic classification. In: Proceedings of the 50th Annual Meeting of the
  Association for Computational Linguistics: Short Papers-Volume 2. pp. 90--94.
  Association for Computational Linguistics (2012)

\bibitem{Wiebe:00}
Wiebe, J.: Learning subjective adjectives from corpora. In: AAAI/IAAI. pp.
  735--740 (2000)

\bibitem{Wilson:05}
Wilson, T., Wiebe, J., Hoffmann, P.: Recognizing contextual polarity in
  phrase-level sentiment analysis. In: Proceedings of the conference on human
  language technology and empirical methods in natural language processing. pp.
  347--354. Association for Computational Linguistics (2005)

\bibitem{Zhang:11}
Zhang, L., Ghosh, R., Dekhil, M., Hsu, M., Liu, B.: Combining lexicon-based and
  learning-based methods for twitter sentiment analysis  (2011)

\end{thebibliography}
\bibliographystyle{cicling}

\end{document}